\documentclass[11pt,a4paper]{article}
\usepackage{acl2015}
\usepackage{times}
\usepackage{url}
\usepackage{latexsym}

\usepackage{caption} 
\usepackage{subcaption} 
\usepackage{multirow} 
\usepackage{amsmath} 
\usepackage{amssymb} 
\usepackage{graphicx}
\usepackage{color} 
\usepackage{pstricks} 
\usepackage{arydshln} 

\newcommand{\imgExt}{eps}
\newcommand{\eq}[1]{Eq.~(\ref{#1})}

\newcommand{\correct}[1]{\textit{\color{blue} #1}}
\newcommand{\wrong}[1]{\textbf{\color{red} #1}}
\newcommand{\edit}[1]{{#1}} %
\newcommand{\hide}[1]{}

\newcommand{\MB}[1]{\mbox{\boldmath{$#1$}}} 
\newcommand{\inR}[1]{\in \mathbb{R}^{#1}} 

\newcommand{\open}[1]{\left(#1\right)} 
\newcommand{\W}[1]{\MB{W_{#1}}} 
\newcommand{\h}[1]{\MB{h}_{#1}} 
\newcommand{\hi}{\MB{h}_{t}} 
\newcommand{\his}{\MB{\bar{h}}_{s}} 
\newcommand{\hs}{\MB{\tilde{h}}_{t}} 
\newcommand{\al}{\MB{a}_{t}} 
\newcommand{\co}{\MB{c}_{t}} 
\newcommand{\pt}{p_{t}} 
\newcommand{\tgt}[1]{y_{#1}} 
\newcommand{\src}[1]{x_{#1}} 
\newcommand{\eos}{$<$\texttt{eos}$>$}
\newcommand{\unk}{$<$\texttt{unk}$>$}
\newcommand{\tp}[1]{#1^\top} 
\DeclareMathOperator{\softmax}{softmax}
\DeclareMathOperator{\sigmoid}{sigmoid}
\DeclareMathOperator{\alignf}{align}
\DeclareMathOperator{\score}{score}

\newcommand{\sotaold}{23.0} 
\newcommand{\sotanew}{25.9} 
\newcommand{\attngain}{5.0} 
\newcommand{\localm}{local-m} 
\newcommand{\localp}{local-p} 

\title{Effective Approaches to Attention-based Neural Machine Translation}

\author{Minh-Thang Luong $\mbox{ }$ $\mbox{ }$ $\mbox{ }$ Hieu Pham $\mbox{ }$ $\mbox{ }$ $\mbox{ }$ Christopher D. Manning\\
  Computer Science Department, Stanford University, Stanford, CA 94305 \\
  {\tt \{lmthang,hyhieu,manning\}@stanford.edu} 
}

\date{}

\begin{document}
\maketitle
\begin{abstract}
An attentional mechanism has lately been used to improve neural machine translation (NMT)
by
selectively focusing on parts of the source sentence during translation. However,
there has been little work exploring useful architectures for attention-based
NMT. This paper examines two simple and effective classes of attentional
mechanism: a {\it global} approach which always attends to all source words and
a {\it local} one that only looks at a subset of source words at a time. 
We demonstrate the effectiveness of both approaches on the WMT translation
tasks between English and German in both directions. With local
attention, we achieve a significant gain of \attngain{} BLEU points over
non-attentional systems that 
already incorporate known techniques such as dropout. Our ensemble 
model using different attention architectures yields a new
state-of-the-art result in the WMT'15 English to German
translation task with \sotanew{} BLEU points, an improvement of 1.0 BLEU points over the existing
best system backed by NMT and an $n$-gram reranker.\footnote{All our code and models
are publicly available at \url{http://nlp.stanford.edu/projects/nmt}.}

\end{abstract}

\section{Introduction}
\label{sec:intro}
\begin{figure}
\centering
\includegraphics[width=0.45\textwidth, clip=true, trim= 0 0 0 0]{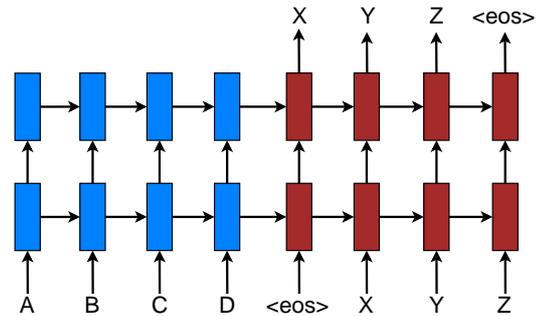} 
\caption{{\bf Neural machine translation} -- a stacking recurrent architecture for translating a source sequence \texttt{A B C D} into a target sequence \texttt{X Y Z}. Here, \eos{} marks the end of a sentence.
} 
\label{f:lstm}
\end{figure}

Neural Machine Translation (NMT) achieved state-of-the-art performances in
large-scale translation tasks such as from English to French \cite{luong15} and
English to German \cite{jean15}. NMT is appealing since it requires minimal
domain knowledge and is conceptually simple. The model by \newcite{luong15} reads through all the source words until the end-of-sentence symbol \eos{} is reached. It then starts emitting one target word at a time, as illustrated in Figure~\ref{f:lstm}. NMT is often a large neural network that is trained in an end-to-end fashion and has the ability to generalize well to very long word sequences. This means the model does not have to explicitly store gigantic phrase tables and language models as in the case of standard MT; hence, NMT has a small memory footprint. Lastly, implementing NMT decoders is easy unlike the highly intricate decoders in standard MT \cite{Koehn:2003:SMT}.

In parallel, the concept of ``attention" has gained popularity recently in
training neural networks, allowing models to learn alignments between different
modalities, e.g., between image objects and agent actions in the dynamic control
problem \cite{mnih14}, between speech frames and text in the speech recognition
task \cite{jan14},  or between visual features of a picture and its text
description in the image caption generation task \cite{xu15}. In the context of
NMT, \newcite{bog15} has successfully applied such attentional mechanism to
jointly translate and align words. To the best of our knowledge, there has not
been any other work exploring the use of attention-based architectures for NMT.

In this work, we design, with simplicity and effectiveness in mind, two novel
types of attention-based models: a {\it global} approach in which all source
words are attended and a {\it local} one whereby only a subset of source words
are considered at a time. The former approach resembles the model of
\cite{bog15} but is simpler architecturally. The latter can be viewed as an
interesting blend between the {\it hard} and {\it soft} attention models
proposed in \cite{xu15}: \edit{it is computationally less expensive than the
global model or the soft attention; at the same time, unlike the hard attention,
the local attention is
differentiable almost everywhere, making it easier to implement and
train.\footnote{There is a recent work by \newcite{draw15}, which is very
similar to our local attention and applied to the image generation task.
However, as we detail later, our model is much simpler and can achieve good performance for NMT.} Besides, we also examine various
alignment functions for our attention-based models.}

Experimentally, we demonstrate that both of our approaches are
effective in the WMT translation tasks between English and German in  both
directions. \edit{Our attentional models yield a boost of up to \attngain{} BLEU over
non-attentional systems which already incorporate known techniques such as
dropout. For English to German translation, we achieve new state-of-the-art
(SOTA)
results for both WMT'14 and WMT'15, outperforming previous SOTA systems, backed by
NMT models and $n$-gram LM rerankers, by more than 1.0 BLEU. We conduct
extensive analysis to evaluate our models in terms of learning, the ability to
handle long sentences, choices of attentional architectures, alignment quality, and translation
outputs. 
}

\section{Neural Machine Translation}
\label{sec:nmt}
A neural machine translation system is a neural network that directly models the conditional probability $p(\tgt{}|\src{})$ of translating
a source sentence, $\src{1},\ldots,\src{n}$, to a target sentence, $\tgt{1},\ldots,\tgt{m}$.\footnote{All sentences are assumed to terminate with a special ``end-of-sentence'' token \eos{}.}
A basic form of NMT consists of two components: (a) an {\it encoder} which computes a representation $\MB{s}$ for each source sentence  and (b) a {\it decoder} which generates one target word at a time and hence decomposes the conditional probability as:
\begin{equation}
\log p(\tgt{}|\src{}) = \sum_{j=1}^m \nolimits \log p\open{\tgt{j}|\tgt{<j},\MB{s}}
\end{equation}

A natural choice to model such a decomposition in the decoder is to use a recurrent neural network (RNN) architecture, which most of the recent NMT work such as \cite{kal13,sutskever14,cho14,bog15,luong15,jean15} have in common. They, however, differ in terms of which RNN architectures are used for the decoder and how the encoder computes the source sentence representation $\MB{s}$.

\newcite{kal13} used an RNN with the standard hidden unit for the decoder and a
convolutional neural network for encoding the source sentence representation. On
the other hand, both \newcite{sutskever14} and \newcite{luong15} stacked
multiple layers of an RNN with a Long Short-Term Memory (LSTM) hidden unit for
both the encoder and the decoder. \newcite{cho14}, \newcite{bog15}, and
\newcite{jean15} all adopted a different version of the RNN with an
LSTM-inspired hidden unit, the gated recurrent unit (GRU), for both
components.\footnote{They all used a single RNN layer except for the latter two
works which utilized a bidirectional RNN for the encoder.}

In more detail, one can parameterize the probability of decoding each word $y_j$ as:
\begin{equation}
p\left(\tgt{j}|\tgt{<j},\MB{s}\right) = \softmax\open{g\open{\h{j}}}
\end{equation}
with $g$ being the transformation function that outputs a vocabulary-sized
vector.\footnote{One can provide $g$ with other inputs such as the currently
predicted word $\tgt{j}$ as in \cite{bog15}.} Here, $\h{j}$ is the RNN hidden
unit, abstractly computed as:
\begin{equation}
\h{j} = f(\h{j-1}, \MB{s}),
\end{equation}
where $f$ computes the current hidden state given the previous hidden state and
can be either a vanilla RNN unit, a GRU, or an LSTM unit. In \cite{kal13,sutskever14,cho14,luong15}, the source representation $\MB{s}$ is only used once to initialize the decoder hidden state. On the other hand, in \cite{bog15,jean15} and this work, $\MB{s}$, in fact, implies a set of source hidden states which are consulted throughout the entire course of the translation process. Such an approach is referred to as an attention mechanism, which we will discuss next.

In this work, following \cite{sutskever14,luong15}, we use the stacking LSTM architecture for our NMT systems, as illustrated in Figure~\ref{f:lstm}.
We use the LSTM unit defined in \cite{zaremba15}. Our training objective is formulated as follows:
\begin{equation}
J_t = \sum_{(\src{},\tgt{}) \in \mathbb{D}} \nolimits -\log p(\tgt{}|\src{})
\label{e:j_t}
\end{equation}
with $\mathbb{D}$ being our parallel training corpus.

\section{Attention-based Models}
\label{sec:attn}
\begin{figure}
\centering
\rput(6.3,6.2){$\tgt{t}$}
\rput(6.6,5.7){$\hs$}
\rput(2.3,3.5){$\co$}
\rput(3.2,2.5){$\al$}
\rput(6.75,1.7){$\hi$}
\rput(5.5,2.8){$p_t$}
\rput(0.2,2.1){$\his$}
\includegraphics[width=0.45\textwidth, clip=true, trim= 0 0 0 0]{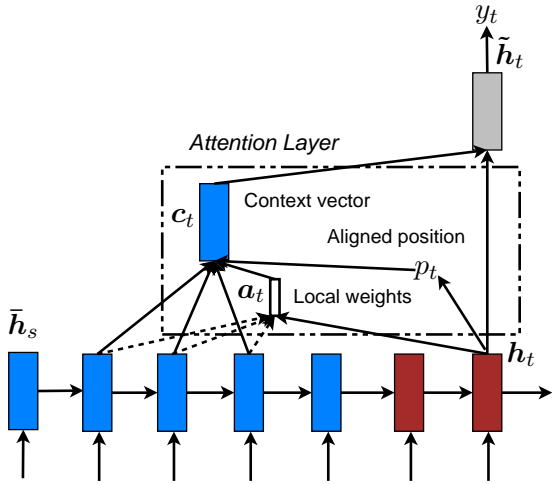} 
\caption{{\bf Local attention model} -- the model first predicts a single
aligned position $p_t$ for the current target word. A window centered around the
source position $p_t$ is then used to compute a context vector $\co$, a weighted
average of the source hidden states in the window. The weights $\MB{\al}$ are
inferred from the current target state $\hi$ and those source states $\his$ in
the window.
} 
\label{f:hard_attn}
\end{figure}

Our various attention-based models are classifed into two broad categories, {\it global} and {\it local}. These classes differ in terms of whether the ``attention'' is placed on all source positions or on only a few source positions. We illustrate these two model types in Figure~\ref{f:soft_attn} and \ref{f:hard_attn} respectively.

Common to these two types of models is the fact that at each time step $t$ in the decoding phase, both approaches first take as input the hidden state $\hi$ at the top layer of a stacking LSTM. The goal is then to derive a context vector $\co$ that captures relevant source-side information to help predict the current target word $\tgt{t}$. While these models differ in how the context vector $\co$ is derived, they share the same subsequent steps. 

Specifically, given the target hidden state $\hi$ and the source-side context vector $\co$, we employ a simple concatenation layer to combine the information from both vectors to produce an attentional hidden state as follows:
\begin{equation}
\hs = \tanh(\W{c}[\co; \hi])
\label{e:hs}
\end{equation} 

The attentional vector $\hs$ is then fed through the softmax layer to produce the predictive distribution formulated as:
\begin{equation}
p(\tgt{t}|\tgt{<t},\src{}) = \softmax(\W{s}\hs)
\label{e:predict}
\end{equation} 

We now detail how each model type computes the source-side context vector $\co$.

\subsection{Global Attention}
\label{subsec:global}
\begin{figure}
\centering
\rput(6.3,6.2){$\tgt{t}$}
\rput(6.6,5.7){$\hs$}
\rput(2.3,3.5){$\co$}
\rput(4.3,2.6){$\al$}
\rput(6.75,1.7){$\hi$}
\rput(0.2,2.1){$\his$}
\includegraphics[width=0.45\textwidth, clip=true, trim= 0 0 0 0]{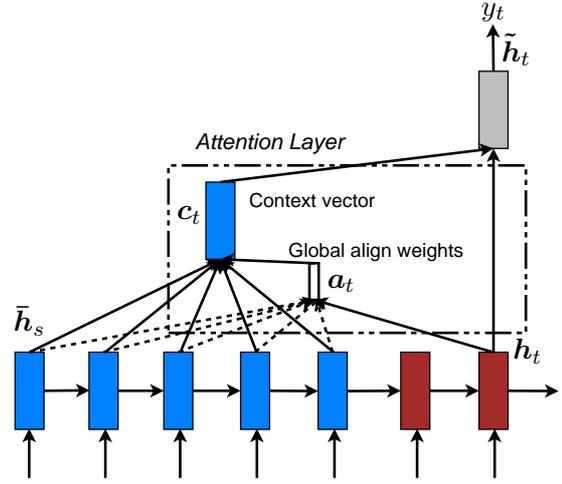} 
\caption{{\bf Global attentional model} -- at each time step $t$, the model infers a {\it variable-length} alignment weight vector $\al$ based on the current target state $\hi$ and all source states $\his$. A global context vector $\co$ is then computed as the weighted average, according to $\al$, over all the source states. 
} 
\label{f:soft_attn}
\end{figure}

The idea of a global attentional model is to consider all the hidden states of
the encoder when deriving the context vector $c_t$. In this model type, a
variable-length alignment vector $\al$, whose size equals the number of time
steps on the source side, is derived by comparing the current target hidden
state $\hi$ with each source hidden state $\his$:
\begin{align}
\label{e:al}
\al(s)&=\alignf(\hi, \his) \\
&=\frac{\exp \open{\score(\hi, \his)}}{\sum_{s'} \exp \open{\score(\hi,
\MB{\bar{h}}_{s'})}} \notag
\end{align}
Here, $\score$ is referred as a {\it content-based} function for which we consider three different
alternatives:
\begin{equation*}
\score(\hi, \his)\!=\!\begin{cases}
    \tp{\hi} \his & \mbox{{\it dot}}\\
    \tp{\hi} \MB{W_a} \his & \mbox{{\it general}} \\
    \tp{\MB{v}_a}\tanh\open{\MB{W_a} [\hi; \his]} & \mbox{{\it concat}}
\end{cases}
\end{equation*}

Besides, in our early attempts to build attention-based models, we use
a {\it location-based} function in which the alignment scores are
computed from solely the target hidden state $\hi$ as
follows:
\begin{equation}
\al = \softmax(\W{a}\hi) \mbox{ } \mbox{ } \mbox{ } \mbox{ } \mbox{ } \mbox{ } \mbox{ } \mbox{ } \mbox{ } \mbox{ } \mbox{ } \mbox{ } \mbox{ } \mbox{ } \mbox{ } \mbox{ } \mbox{{\it location}}
\label{e:location}
\end{equation}
Given the alignment vector as weights, the
context vector $c_t$ is computed as the  weighted average over all the source hidden states.\footnote{\edit{\eq{e:location} implies that
all alignment vectors $\al$ are of the same length. For short sentences, we only
use the top part of $\al$ and for long sentences, we ignore words near the end.}}

\textit{Comparison to \cite{bog15}} --
While our global attention approach is similar in spirit to the model proposed
by \newcite{bog15}, there are several key differences which reflect how we have
both simplified and generalized from the original model. First, we simply use
hidden states at the top LSTM layers in both the encoder and decoder as
illustrated in Figure~\ref{f:soft_attn}. \newcite{bog15}, on the other hand,
use the concatenation of the forward and backward source 
hidden states in the bi-directional encoder and target hidden
states in their non-stacking uni-directional decoder. Second, our computation
path is simpler; we go from $\hi \rightarrow \al \rightarrow \co \rightarrow
\hs$ then make a prediction as detailed in \eq{e:hs}, \eq{e:predict}, and
Figure~\ref{f:soft_attn}. On the other hand, at any time $t$, \newcite{bog15} build from the previous hidden state $\h{t-1} \rightarrow \al \rightarrow \co \rightarrow \hi$, which, in turn, goes through a deep-output and a maxout layer before making predictions.\footnote{We will refer to this difference again in Section~\ref{subsec:input}.} Lastly, \newcite{bog15} only experimented with one alignment function, the {\it concat} product; whereas we show later that the other alternatives are better.

\subsection{Local Attention}

The global attention has a drawback that it has to attend to all words on the
source side for each target word, which is expensive and can potentially render it impractical to
translate longer sequences, e.g., paragraphs or documents.
To address this deficiency, we propose a {\it local} attentional mechanism that
chooses to focus only on a small subset of the source positions per target word.

This model takes inspiration from the tradeoff between the {\it soft} and {\it
hard} attentional models proposed by \newcite{xu15} to tackle the image caption
generation task. In their work, soft attention refers to the global attention
approach in which weights are placed ``softly'' over all patches in the source
image. The hard attention, on the other hand, selects one patch
of the image to attend to at a time. While less expensive at inference time, the
hard attention model is non-differentiable and requires more complicated
techniques such as variance reduction or reinforcement learning to train.

Our local attention mechanism selectively focuses on a small window of
context and is differentiable. This approach has an advantage of avoiding the expensive computation incurred in
the soft attention and at the same time, is easier to train than the hard
attention approach.
In concrete details, the model first generates an aligned position $p_t$ for each target word at time $t$. The
context vector $\co$ is then derived as a weighted average over the set of source hidden states within the window $[p_t-D, p_t+D]$; $D$ is
empirically selected.\footnote{If the window crosses the sentence boundaries, we
simply ignore the outside part and consider words in the window.} Unlike the global approach, the local alignment vector $\al$ is now fixed-dimensional, i.e., $\inR{2D+1}$. 
We consider two variants of the model as below.

\textit{Monotonic} alignment ({\bf \localm{}}) -- we simply set %
$\pt\!=\!t$ assuming that source and target sequences are roughly
monotonically aligned. The alignment vector $\al$ is defined according to
\eq{e:al}.\footnote{\edit{{\it local-m} is the same as
the global model except that the vector $\al$ is
fixed-length and shorter.}} 

\textit{Predictive} alignment ({\bf \localp{}}) --  %
instead of assuming monotonic alignments, our model predicts an aligned position as follows:
\begin{equation}
\pt = S \cdot \sigmoid(\tp{\MB{v}_p}\tanh(\W{p}\hi)),
\label{e:p}
\end{equation}
$\W{p}$ and $\MB{v}_p$ are the model parameters which will be learned
to predict positions. $S$ is the source sentence length. As a result of $\sigmoid$, $\pt
\in [0, S]$. To favor alignment points near $\pt$, we place a Gaussian distribution centered around $\pt$ . Specifically, our alignment weights are now
defined as:
\begin{equation}
\al(s) = \alignf(\hi, \his) \exp \open{-\frac{(s-\pt)^2}{2\sigma^2}} 
\label{e:align_p}
\end{equation}
We use the same $\alignf$ function as in
\eq{e:al} and the standard deviation is empirically set as
$\sigma\!=\!\frac{D}{2}$. Note that $\pt$ is a {\it real} nummber; whereas $s$
is an {\it integer} within the window centered at $\pt$.\footnote{{\it local-p} is similar to the
local-m model except that we dynamically
compute $\pt$ and use a truncated Gaussian distribution to modify the original alignment
weights $\alignf(\hi, \his)$ as shown in \eq{e:align_p}. By utilizing $\pt$
to derive $\al$, we can compute backprop gradients for $\W{p}$ and $\MB{v}_p$.
This model is differentiable almost everywhere.} 

\textit{Comparison to \cite{draw15}} --
have proposed a {\it selective attention} mechanism, very
similar to our local attention, for the image generation task. Their approach 
allows the model to select an image patch of varying location and zoom. We,
instead, use the same ``zoom'' for all target positions, which greatly
simplifies the formulation and still achieves good
performance.

\subsection{Input-feeding Approach}
\label{subsec:input}
\begin{figure}
\centering
\rput(3,5.6){$\hs$}
\includegraphics[width=0.4\textwidth, clip=true, trim= 0 0 0 0]{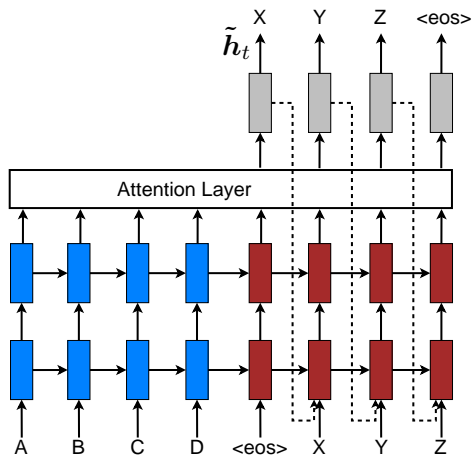} 
\caption{{\bf Input-feeding approach} -- Attentional vectors $\hs$ are fed as inputs to the next time steps to inform the model about past alignment decisions.
} 
\label{f:input}
\end{figure}

In our proposed global and local approaches, the attentional decisions are made
independently, which is suboptimal. Whereas, in standard MT, a {\it coverage}
set is often maintained during the translation process to keep track of which
source words have been translated. Likewise, in attentional NMTs, alignment
decisions should be made jointly taking into account past alignment information.
To address that, we propose an {\it input-feeding} approach in which attentional
vectors $\hs$ are concatenated with inputs at the next time steps as illustrated in
Figure~\ref{f:input}.\footnote{If $n$ is the number of LSTM cells, the
input size of the first LSTM layer is $2n$; those of subsequent
layers are $n$.} The effects of having such connections are two-fold:
(a) we hope to make the model fully aware of previous alignment choices and (b)
we create a very deep network spanning both horizontally and vertically.

{\it Comparison to other work} -- \newcite{bog15}
use context vectors, similar to
our $\co$, in building subsequent hidden states, which can also 
achieve the ``coverage'' effect. However, there has not been any analysis of 
whether such connections are useful as done in this work. Also,
our approach is more general; as illustrated in Figure~\ref{f:input}, it can be
applied to general stacking recurrent architectures, including non-attentional
models.

\newcite{xu15} propose a {\it doubly attentional} approach with an
additional constraint added to the training objective to make sure the model
pays equal attention to all parts of the image during the caption generation process. Such a constraint can also be useful to capture the coverage set effect in NMT that we mentioned earlier. However, we chose to use the input-feeding approach since it provides flexibility for the model to decide on any attentional constraints it deems suitable.

\section{Experiments}
\label{sec:exp}
We evaluate the effectiveness of our models on the WMT translation tasks between
English and German in both directions. newstest2013 (3000 sentences) is used as
a development set to select our hyperparameters. Translation performances are
reported in case-sensitive BLEU \cite{Papineni02bleu} on newstest2014 (2737 sentences) and
newstest2015 (2169 sentences). \edit{Following \cite{luong15}, we report
translation quality using two types of BLEU: (a) {\it
tokenized}\footnote{All texts are tokenized with \texttt{tokenizer.perl} and BLEU
scores are computed with \texttt{multi-bleu.perl}.} BLEU to be comparable with
existing NMT work and (b) {\it NIST}\footnote{With the \texttt{mteval-v13a}
script as per WMT guideline.} BLEU to be comparable
with WMT results.}

\begin{table*}[tbh!]
\centering
\resizebox{15.8cm}{!}{
\begin{tabular}{l|r|c}
\bf{System} & \bf{Ppl} & \bf{BLEU}\\
  \hline
Winning WMT'14 system -- {\it phrase-based + large LM} \cite{buck14} &  & 20.7\\
  \hline
\multicolumn{3}{l}{{\it Existing NMT systems}}\\
  \hline
RNNsearch \cite{jean15} &  & 16.5\\
RNNsearch + unk replace \cite{jean15} &  & 19.0\\
RNNsearch + unk replace + large vocab + {\it ensemble} 8 models \cite{jean15} &  & {\bf 21.6}\\
  \hline
\multicolumn{3}{l}{{\it Our NMT systems}}\\
  \hline
Base & 10.6 & 11.3\\
Base + reverse & 9.9 & 12.6 ({\it +1.3})\\
Base + reverse + dropout & 8.1 & 14.0 ({\it +1.4})\\
  \hdashline
Base + reverse + dropout + global attention ({\it location}) & 7.3 & 16.8 ({\it +2.8}) \\
Base + reverse + dropout + global attention ({\it location}) + feed input & 6.4
& 18.1 ({\it +1.3}) \\
  \hdashline
Base + reverse + dropout + local-p attention ({\it general}) + feed input
& \multirow{ 2}{*}{5.9} & 19.0 ({\it +0.9}) \\
Base + reverse + dropout + local-p attention ({\it general}) + feed input + unk replace
&  & 20.9 ({\it +1.9}) \\
  \hdashline
{\it Ensemble} 8 models + unk replace &  & {\bf 23.0 ({\it +2.1})} \\
\end{tabular}
}
\caption{{\bf WMT'14 English-German results} -- shown are
the perplexities (ppl) and the {\it tokenized} BLEU scores of various systems on newstest2014. We highlight the {\bf
best} system in bold and give {\it progressive} improvements in italic between
consecutive systems. {\it local-p} referes to the local attention with 
predictive alignments. We indicate for each attention model the
alignment score function used in pararentheses. 
}
\label{t:ende}
\end{table*}

\subsection{Training Details}
All our models are trained on the WMT'14 training data consisting of 4.5M
sentences pairs (116M English words, 110M German words). Similar to \cite{jean15}, we limit our vocabularies to be the top 50K most frequent words for both languages. Words not in these shortlisted vocabularies are converted into a universal token \unk{}. 


When training our NMT systems, following \cite{bog15,jean15}, we filter out
sentence pairs whose lengths exceed 50 words and shuffle mini-batches as we
proceed. Our stacking LSTM models have 4 layers, each with 1000 cells, and
1000-dimensional embeddings. We follow \cite{sutskever14,luong15} in training
NMT with similar settings: (a) our parameters are uniformly initialized in
$[-0.1, 0.1]$, (b) we train for 10 epochs using plain SGD, (c) a simple learning
rate schedule is employed -- we start with a learning rate of 1; after 5 epochs,
we begin to halve the learning rate every epoch, (d) our mini-batch size is 128,
and (e) the normalized gradient is rescaled whenever its norm exceeds 5.
Additionally, we also use dropout with probability $0.2$ for our LSTMs as suggested by
\cite{zaremba15}. For dropout models, we train for 12 epochs and start halving
the learning rate after 8 epochs. For local
attention models, we empirically set the window size $D=10$.

Our code is implemented in MATLAB. 
When running on a single GPU device Tesla K40, we achieve a speed of 1K {\it
target} words per second. It takes 7--10 days to completely train a model.

\subsection{English-German Results}
We compare our NMT systems in the English-German task with various other
systems. These include the winning system in WMT'14
\cite{buck14}, a phrase-based system whose language models were trained on a
huge monolingual text, the Common Crawl corpus. For end-to-end NMT systems, to the best of our knowledge, \cite{jean15} is the only work experimenting with this language pair and currently the SOTA system.
We only present results for some of our attention models and will later
analyze the rest in Section~\ref{sec:analysis}. 

As shown in Table~\ref{t:ende}, we achieve progressive improvements when
(a) reversing the source sentence, +$1.3$ BLEU, as proposed in \cite{sutskever14}
and (b) using dropout, +$1.4$ BLEU. On top of that, (c) the global
attention approach gives a significant boost of +$2.8$ BLEU, making 
 our model slightly better than the base attentional system of
 \newcite{bog15} (row {\it RNNSearch}). When (d) using the {\it input-feeding}
approach, we seize another notable gain of +$1.3$ BLEU and outperform their
system. The local attention model with predictive alignments (row {\it \localp}) proves
to be even better, giving us a further improvement of +$0.9$ BLEU on top of the
global attention model. 
It is interesting to observe the trend previously reported in
\cite{luong15} that perplexity strongly correlates with translation quality.
In total, we achieve a significant gain of
\attngain{} BLEU points over the non-attentional baseline, which already includes
known techniques such as source reversing and dropout.

The unknown replacement technique proposed in \cite{luong15,jean15} yields another nice
gain of +$1.9$ BLEU, demonstrating that our attentional models
do learn useful alignments for unknown works. Finally, by ensembling 8 different
models of various settings, e.g., using different attention approaches, with
and without dropout etc., we were able to achieve a {\it new SOTA} result of
$\sotaold{}$
BLEU, outperforming the existing best system \cite{jean15} by +$1.4$ BLEU.

\begin{table}[tbh!]
\centering
\resizebox{8cm}{!}{
\begin{tabular}{l|c}
\bf{System} & \bf{BLEU}\\
  \hline
Top -- {\it NMT + 5-gram rerank} (Montreal) & 24.9 \\
  \hline
Our ensemble 8 models + unk replace & {\bf 25.9} \\
\end{tabular}
}
\caption{{\bf WMT'15 English-German results} -- {\it NIST} BLEU scores of the
winning entry in WMT'15 and our best one on newstest2015.}
\label{t:wmt15ende}
\end{table}

{\it Latest results in WMT'15} -- despite the fact that our models were trained
on WMT'14 with slightly less data, we test them on newstest2015 to demonstrate
that they can generalize well to different test sets. As shown in Table~\ref{t:wmt15ende}, our best
system establishes a {\it new SOTA} performance of $\sotanew{}$ BLEU,
outperforming the existing best system backed by NMT and a 5-gram LM reranker
by +$1.0$
BLEU.

\subsection{German-English Results}
We carry out a similar set of experiments for the WMT'15 translation task from German
to English. 
While our systems have not yet matched the performance of the 
SOTA system, we nevertheless show the effectiveness of our
approaches with large and progressive gains in terms of BLEU as illustrated in
Table~\ref{t:deen}. 
The {\it attentional} mechanism gives us +$2.2$ BLEU gain and on top of that, we
obtain another boost of up to +$1.0$ BLEU from the {\it input-feeding} approach.
Using a better alignment function, the content-based {\it dot} product one,
together with {\it dropout} yields another gain of +$2.7$ BLEU. Lastly, when
applying the unknown word replacement technique, we seize an additional +$2.1$
BLEU, demonstrating the usefulness of attention in aligning rare words.

\section{Analysis}
\label{sec:analysis}
\edit{We conduct extensive analysis to better understand our models in terms
of learning, the ability to handle long sentences, 
choices of attentional architectures, and alignment quality.} All results
reported here are on English-German newstest2014.
\begin{table}
\centering
\resizebox{8cm}{!}{
\begin{tabular}{l|r|c}
\bf{System} & \bf{Ppl.} & \bf{BLEU}\\
  \hline
\multicolumn{3}{l}{{\it WMT'15 systems}}\\
  \hline
SOTA -- {\it phrase-based} (Edinburgh) &  & {\bf 29.2}\\ 
NMT + 5-gram rerank (MILA) &  & 27.6\\ 
  \hline
\multicolumn{3}{l}{{\it Our NMT systems}}\\
  \hline
Base (reverse) & 14.3 & 16.9\\
  \hdashline
+ global ({\it location}) & 12.7 & 19.1 ({\it +2.2}) \\
+ global ({\it location}) + feed & 10.9 & 20.1 ({\it +1.0})\\
  \hdashline
+ global ({\it dot}) + drop + feed & \multirow{ 2}{*}{9.7} & 22.8 ({\it +2.7})\\
+ global ({\it dot}) + drop + feed + unk &  & 24.9 ({\it +2.1})\\
\end{tabular}
}
\caption{{\bf WMT'15 German-English results} -- 
performances of various systems (similar to 
Table~\ref{t:ende}). The {\it base} system already includes source reversing
on which we add {\it global} attention, {\it drop}out, input {\it feed}ing, and
{\it unk} replacement.}
\label{t:deen}
\end{table}

\subsection{Learning curves}
\begin{figure}
\centering
\includegraphics[width=0.48\textwidth, clip=true, trim=140 0 70 0]{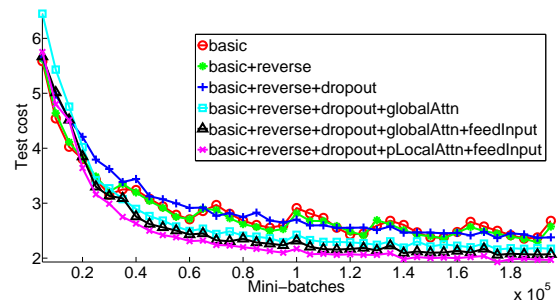} 
\caption{{\bf Learning curves} -- test cost ($\ln$ perplexity) on newstest2014 for English-German NMTs as training progresses.
} 
\label{f:learn}
\end{figure}

We compare models built on top of one another 
as listed in Table~\ref{t:ende}. It is
pleasant to observe in Figure~\ref{f:learn} a clear separation between non-attentional and attentional
models. The input-feeding approach and the local attention
model also demonstrate their abilities in driving the test costs lower. The
non-attentional model with 
dropout (the blue + curve) learns slower than other non-dropout models, but
as time goes by, it becomes more robust in terms of minimizing test errors.

\subsection{Effects of Translating Long Sentences}
\begin{figure}[tbh!]
\centering
\includegraphics[width=0.5\textwidth, clip=true, trim=120 0 70 0]{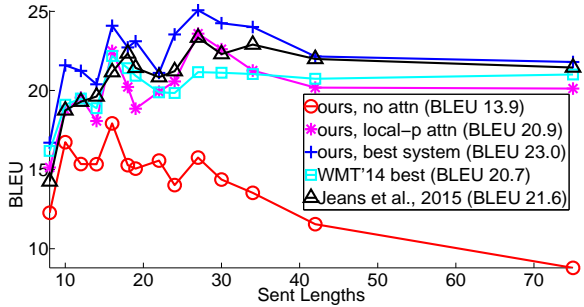} 
\caption{{\bf Length Analysis} -- translation qualities of different
systems as sentences become longer.
} 
\label{f:length}
\end{figure}

We follow \cite{bog15} to group sentences of similar lengths together and
compute a BLEU score per group. Figure~\ref{f:length} shows that
our attentional models are more effective than the non-attentional one in
handling long sentences: the quality does not degrade as sentences
become longer. Our best model (the blue + curve) outperforms all other systems in all length buckets.

\subsection{Choices of Attentional Architectures}
We examine different attention models ({\it global, local-m, local-p}) and different
alignment functions ({\it location, dot, general, concat}) as described in
Section~\ref{sec:attn}. Due to limited
resources, we cannot run all the possible combinations.
However, results in Table~\ref{t:attnChoices} do give us some idea about
different choices. 
The {\it location-based} function does not learn good
alignments: the {\it global (location)} model can only obtain a small
gain when performing unknown word replacement compared to using other alignment
functions.\footnote{There is a subtle difference in how we retrieve alignments
for the different alignment functions. At time step $t$ in which we receive
$\tgt{t-1}$ as input and then compute $\hi, \al, \co$, and $\hs$ before
predicting $\tgt{t}$, the alignment vector $\al$ is used as alignment
weights for (a) the predicted word $\tgt{t}$ in the {\it location-based}
alignment functions and (b) the input word $\tgt{t-1}$ in the {\it content-based}
functions.}
For {\it content-based} functions, our implementation {\it concat} does not yield good performances
and more analysis should be done to understand the
reason.\footnote{With {\it concat}, the perplexities achieved by different models are 6.7 (global), 7.1
(local-m), and 7.1 (local-p). Such high perplexities could be due to the fact
that we simplify the matrix \MB{W_a} to set the part that corresponds to $\his$
to identity.} It is interesting to observe that {\it dot} works
well for the global attention and {\it general} is better for the local
attention.
Among the different models, the local attention model with predictive alignments ({\it
local-p}) is best, both in terms of perplexities and BLEU.

\begin{table}
\centering
\resizebox{8cm}{!}{
\begin{tabular}{l|r|c|c}
\multirow{ 2}{*}{\bf{System}} & \multirow{ 2}{*}{\bf{Ppl}} &
\multicolumn{2}{c}{{\bf BLEU}}\\
\cline{3-4}
& & Before & After unk \\
  \hline
global (location) & 6.4 & 18.1 & 19.3 (+1.2) \\
global (dot) & 6.1 & 18.6 & 20.5 (+1.9) \\
global (general) & 6.1 & 17.3 & 19.1 (+1.8) \\
  \hline
local-m (dot) & $>$7.0 & x & x \\
local-m (general) & 6.2 & 18.6 & 20.4 (+1.8) \\
  \hline
local-p (dot) & 6.6 & 18.0 & 19.6 (+1.9) \\
local-p (general) & {\bf 5.9} & {\bf 19} & {\bf 20.9 (+1.9)} \\
\end{tabular}
}
\caption{{\bf Attentional Architectures} -- performances of different
attentional
models. We trained two local-m (dot) models; both have
ppl $>7.0$.}
\label{t:attnChoices}
\end{table}

\begin{table*}[tbh!]
\centering
\resizebox{15cm}{!}{
\begin{tabular}{c|p{15cm}}
\multicolumn{2}{l}{{\bf English-German translations}}\\
  \hline
src & Orlando Bloom and Miranda Kerr still love each other \\
  \hline
ref & Orlando Bloom und \correct{Miranda Kerr} lieben sich noch immer \\
  \hline
{\it best} & Orlando Bloom und \correct{Miranda Kerr} lieben einander noch immer . \\
  \hline
base & Orlando Bloom und \wrong{Lucas Miranda} lieben einander noch immer .\\
  \hline
  \hline
src & $''$ We $'$ re pleased the FAA recognizes that an enjoyable passenger experience is not incompatible with safety and security , $''$ said Roger Dow , CEO of the U.S. Travel Association . \\
  \hline
ref & `` Wir freuen uns , dass die FAA erkennt , dass ein angenehmes
Passagiererlebnis nicht im Widerspruch zur Sicherheit steht '' , sagte
\correct{Roger Dow} , CEO der U.S. Travel Association . \\
  \hline
{\it best} & $''$ Wir freuen uns , dass die FAA anerkennt , dass ein angenehmes ist
nicht mit Sicherheit und Sicherheit \correct{unvereinbar} ist $''$ , sagte \correct{Roger
Dow} , CEO der US - die . \\
  \hline
base & $''$ Wir freuen uns \"{u}ber die $<$unk$>$ , dass ein $<$unk$>$ $<$unk$>$
mit Sicherheit nicht \wrong{vereinbar} ist mit Sicherheit und Sicherheit $''$ ,
sagte \correct{Roger} \wrong{Cameron} , CEO der US - $<$unk$>$ . \\
  \hline
  \hline
\multicolumn{2}{l}{{\bf German-English translations}}\\
  \hline
src & In einem Interview sagte Bloom jedoch , dass er und Kerr sich noch immer lieben . \\
  \hline
ref & However , in an interview , Bloom has said that he and \correct{Kerr} still love each other . \\
  \hline
{\it best} & In an interview , however , Bloom said that he and \correct{Kerr} still love . \\
  \hline
base & However , in an interview , Bloom said that he and \wrong{Tina} were still $<$unk$>$ . \\
  \hline
  \hline
src & Wegen der von Berlin und der Europ\"{a}ischen Zentralbank verh\"{a}ngten strengen Sparpolitik in Verbindung mit der Zwangsjacke , in die die jeweilige nationale Wirtschaft durch das Festhalten an der gemeinsamen W\"{a}hrung gen\"{o}tigt wird , sind viele Menschen der Ansicht , das Projekt Europa sei zu weit gegangen \\ 
  \hline
ref & The \correct{austerity imposed by Berlin and the European Central Bank , coupled with the straitjacket} imposed on national economies through adherence to the common currency , has led many people to think Project Europe has gone too far .\\
  \hline
{\it best} & Because of the strict \correct{austerity measures imposed by Berlin
and the European Central Bank in connection with the straitjacket} in which the
respective national economy is forced to adhere to the common currency , many
people believe that the European project has gone too far . \\
  \hline
base & Because of the pressure \wrong{imposed by the European Central Bank and the Federal Central Bank with the strict austerity} imposed on the national economy in the face of the single currency , many people believe that the European project has gone too far .\\
\end{tabular}
}
\caption{{\bf Sample translations} -- 
for each example, we show the source ({\it src}), the human translation ({\it
ref}), the translation from our best model ({\it best}), and the
translation of a non-attentional model ({\it base}).  We italicize some
\correct{correct} translation segments and highlight a few \wrong{wrong} ones in
bold.} 
\label{t:sample}
\end{table*}

\subsection{Alignment Quality}
A by-product of attentional models are word alignments. While \cite{bog15}
visualized alignments for some sample sentences and 
observed gains in translation quality as an indication of a working attention
model, no work has assessed the alignments learned as a whole. In contrast, we
set out to evaluate the alignment quality using the alignment error rate (AER)
metric.

\begin{table}
  \begin{center}
    \begin{tabular}{c c}
      {\bf Method} & {\bf AER} \\
      \hline
      global (location) & $0.39$ \\
      local-m (general)  & $0.34$ \\
      local-p (general) & $0.36$ \\
      \hdashline
      ensemble & $0.34$ \\
      \hline
      Berkeley Aligner & $0.32$ \\
    \end{tabular}
  \end{center}
  \caption{{\bf AER scores} -- results of various models on the RWTH
  English-German alignment data.}
  \label{t:alignment}
\end{table}

Given the gold alignment data provided by RWTH for 508 English-German
Europarl sentences, we ``force'' decode our attentional models to
produce translations that match the references. We extract only one-to-one
alignments by selecting the source word with the highest alignment
weight per target word. Nevertheless, as shown in Table~\ref{t:alignment}, we were able to achieve AER scores
comparable to the one-to-many alignments obtained by the Berkeley aligner
\cite{liang06alignment}.\footnote{We concatenate the 508 sentence pairs with 1M
sentence pairs from WMT and run the Berkeley aligner.}

We also found that the alignments produced by local attention models achieve
lower AERs than those of the global one. The AER obtained by the ensemble, while
good, is not better than the local-m AER, suggesting the well-known
observation that AER and translation scores are not well correlated \cite{fraser07}.
We show some alignment visualizations in Appendix~\ref{sec:visual}.

\subsection{Sample Translations}
\label{sec:sample}
We show in Table~\ref{t:sample} sample translations in both directions. It it
appealing to observe the effect of attentional models in correctly translating
names such as ``Miranda Kerr'' and ``Roger Dow''. Non-attentional models, while producing sensible names from a language
model perspective, lack the direct connections from the source side to make
correct translations. 
We also observed an interesting case in the second
example, which requires translating the {\it doubly-negated} phrase, ``not incompatible''.
The attentional model correctly produces ``nicht $\dots$ unvereinbar'';
whereas the non-attentional model generates ``nicht vereinbar'', meaning
``not compatible''.\footnote{The reference uses a more fancy translation of
``incompatible'', which is ``im Widerspruch zu etwas stehen''. Both models, however, failed to translate ``passenger
experience''.} The attentional
model also demonstrates its superiority in translating long sentences as in
the last example.


\section{Conclusion}
\label{sec:conclude}
In this paper, we propose two simple and effective attentional mechanisms for
neural machine translation: the {\it global} approach which always looks at all
source positions and the {\it local} one that only attends to a subset of source
positions at a time. We test the effectiveness of our models in the WMT
translation tasks between English and German in both directions. 
Our local attention yields large gains of up to
$\attngain{}$ BLEU over non-attentional models which already incorporate known
techniques such as dropout. For the English to German translation direction, our
ensemble model has established new state-of-the-art
results for both WMT'14 and WMT'15, outperforming existing
best systems, backed by NMT models and $n$-gram LM rerankers, by more than 1.0 BLEU.

We have compared various alignment functions and shed light on which functions
are best for which attentional models.
Our analysis shows that attention-based NMT models are superior to
non-attentional ones in many cases, for example in translating names and
handling long
sentences.

\section*{Acknowledgment}
We gratefully acknowledge support from a gift from Bloomberg L.P.
and the support of NVIDIA Corporation with the donation of Tesla K40 GPUs.
We thank Andrew Ng and his group as well as the Stanford Research
Computing for letting us use their computing
resources.
We thank Russell Stewart for helpful discussions on the 
models. Lastly, we thank Quoc Le, Ilya Sutskever, Oriol Vinyals,
Richard Socher, Michael Kayser, Jiwei Li, Panupong Pasupat, Kelvin
Guu, members of the Stanford NLP Group and the annonymous reviewers for their valuable comments and feedback.

\bibliography{emnlp15}
\bibliographystyle{acl2015}
\appendix
\section{Alignment Visualization}
\label{sec:visual}
\begin{figure*}
  \begin{center}
    \includegraphics[scale=0.33]{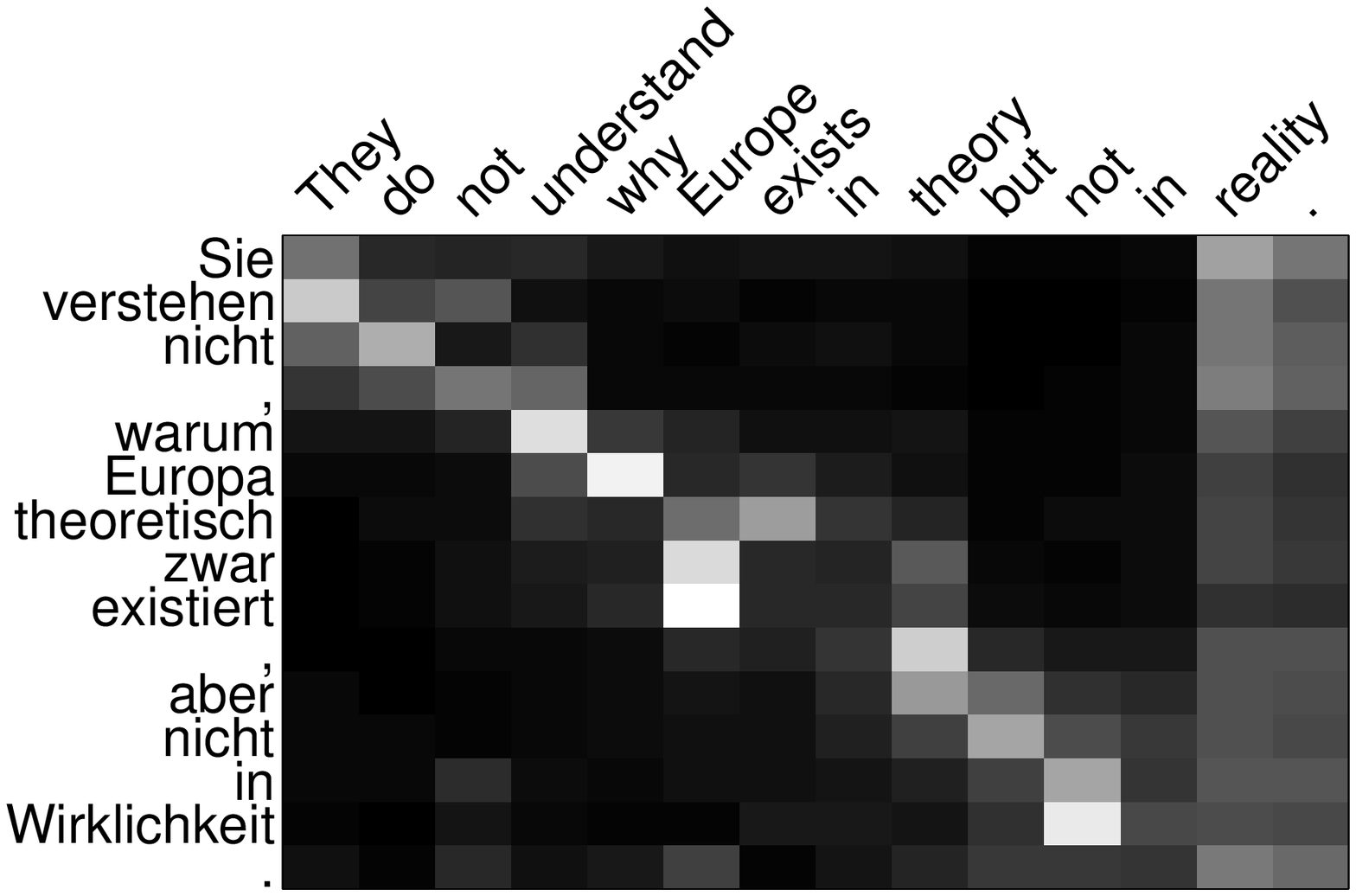}
    \includegraphics[scale=0.33]{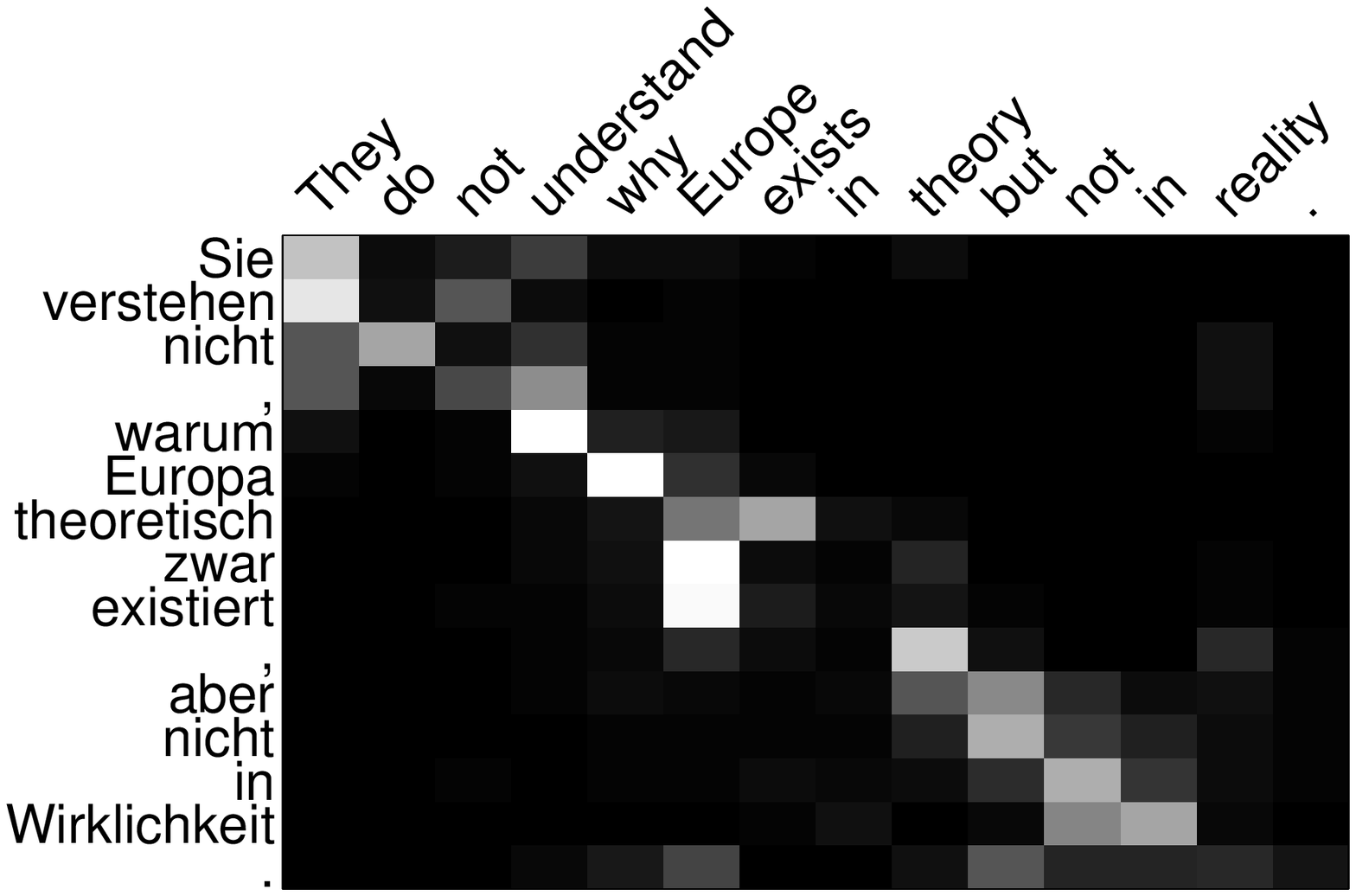}
    \includegraphics[scale=0.33]{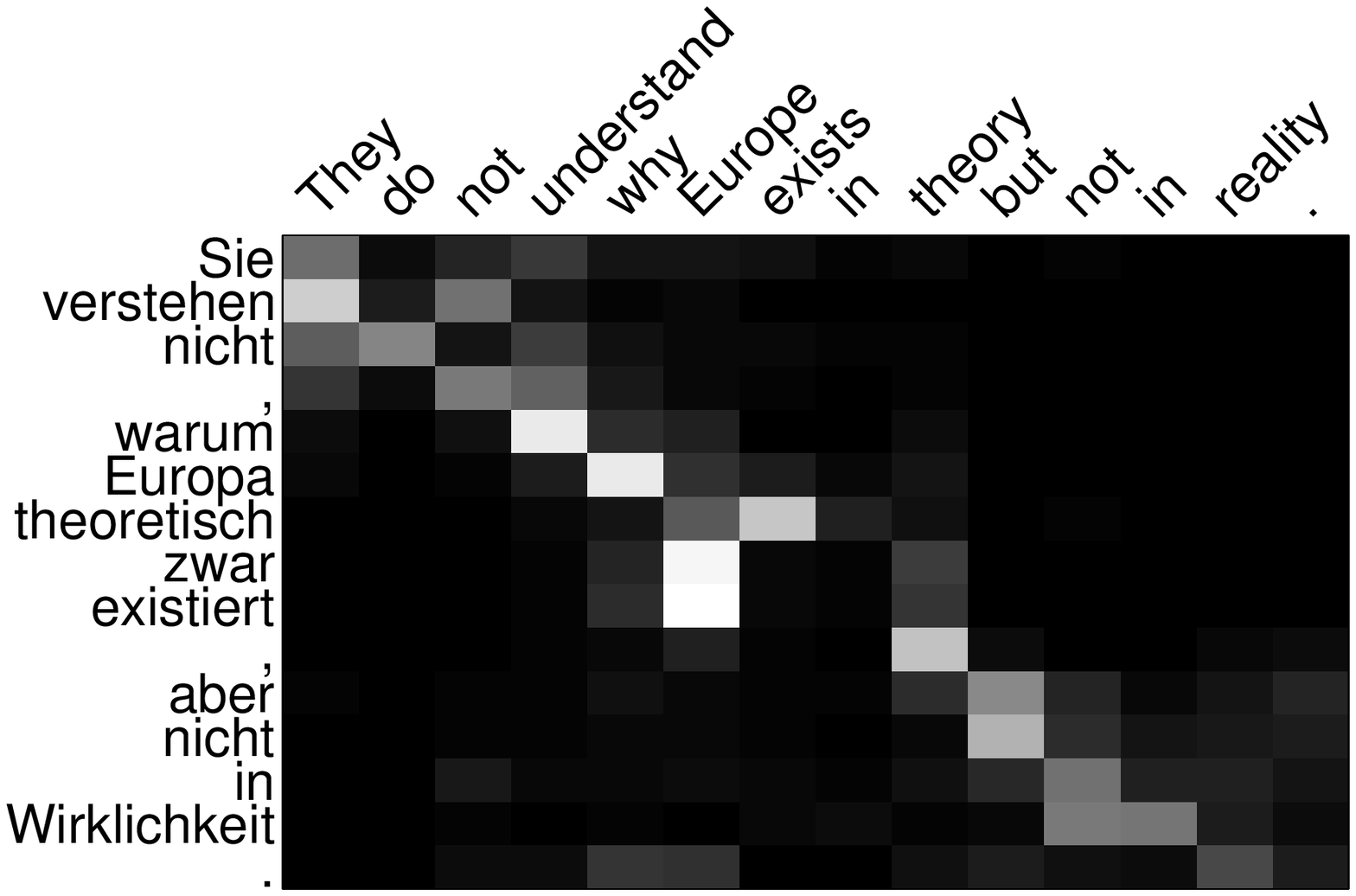}
    \includegraphics[scale=0.33]{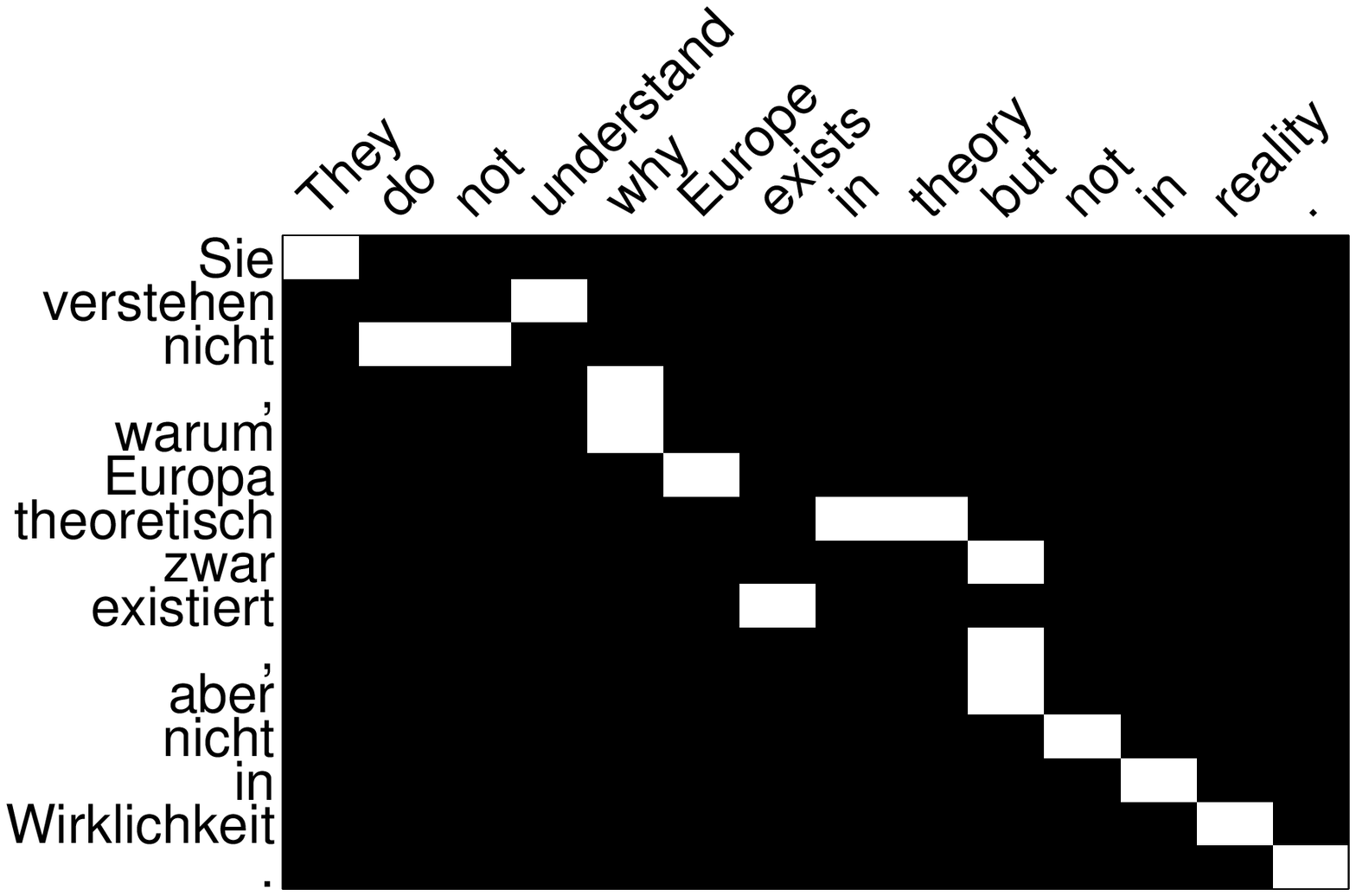}
  \end{center}
  \caption{{\bf Alignment visualizations} -- shown are images of the attention
  weights learned by various models: (top left) global, (top right)
  local-m, and (bottom left) local-p. The {\it gold} alignments are
  displayed at the bottom right corner.
  }
  \label{i:alignment}
\end{figure*}

We visualize the alignment weights produced by our different attention
models in Figure~\ref{i:alignment}. The visualization of the local
attention model is much sharper than that of the global one. This contrast matches
our expectation that local attention is designed to only focus on a subset of
words each time. Also, since we
translate from English to German and reverse the source English sentence, the white strides
at the words {\it ``reality''} and {\it ``.''} in the global attention model reveals an
interesting access pattern: it tends to refer back to the beginning of the
source sequence. 

Compared to the alignment visualizations in \cite{bog15}, our
alignment patterns are not as sharp as theirs. Such difference could possibly be due to
the fact that translating from English to German is
harder than translating into French as done in \cite{bog15},
which is an interesting point to examine in future work.


\end{document}